\documentclass{article}

\usepackage{PRIMEarxiv}

\usepackage[utf8]{inputenc} 
\usepackage[T1]{fontenc}    
\usepackage{hyperref}       
\usepackage{url}            
\usepackage{booktabs}       
\usepackage{amsfonts}       
\usepackage{nicefrac}       
\usepackage{microtype}      
\usepackage{lipsum}
\usepackage{fancyhdr}       
\usepackage{graphicx}       
\graphicspath{{media/}}     

\usepackage{amsmath}
\usepackage{graphicx}
\usepackage{amsmath}
\newtheorem{theorem}{Theorem}

\usepackage{algorithm}
\usepackage{algorithmic}
\usepackage{booktabs} 
\usepackage{multirow}
\usepackage{amsmath}
\usepackage{amssymb}
\usepackage{pgfplots}
\usepackage{authblk}
\pgfplotsset{compat=1.18}
\title{KA-GNN: Kolmogorov-Arnold Graph Neural Networks for Molecular Property Prediction
}

\author[1,2,3]{Longlong Li}
\author[3]{Yipeng Zhang}
\author[1]{Guanghui Wang}
\author[3]{Kelin Xia}

\affil[1]{School of Mathematics, Shandong University, Jinan 250100, China}
\affil[2]{Data Science Institute, Shandong University, Jinan 250100, China}
\affil[3]{Division of Mathematical Sciences, School of Physical and Mathematical Sciences, Nanyang Technological University, Singapore 637371, Singapore}

\affil[ ]{Emails: longlee@mail.sdu.edu.cn, yipeng001@e.ntu.edu.sg, ghwang@sdu.edu.cn, xiakelin@ntu.edu.sg} 

\date{}

\begin{document}
\maketitle

\begin{abstract}
As key models in geometric deep learning, graph neural networks have demonstrated enormous power in molecular data analysis. Recently, a specially-designed learning scheme, known as Kolmogorov-Arnold Network (KAN), shows unique potential for the improvement of model accuracy, efficiency, and explainability. Here we propose the first non-trivial Kolmogorov-Arnold Network-based Graph Neural Networks (KA-GNNs), including KAN-based graph convolutional networks(KA-GCN) and KAN-based graph attention network (KA-GAT). The essential idea is to utilizes KAN's unique power to optimize GNN architectures at three major levels, including node embedding, message passing, and readout. Further, with the strong approximation capability of Fourier series, we develop Fourier series-based KAN model and provide a rigorous mathematical prove of the robust approximation capability of this Fourier KAN architecture. To validate our KA-GNNs, we consider seven most-widely-used benchmark datasets for molecular property prediction and extensively compare with existing state-of-the-art models. It has been found that our KA-GNNs can outperform traditional GNN models. More importantly, our Fourier KAN module can not only increase the model accuracy but also reduce the computational time. This work not only highlights the great power of KA-GNNs in molecular property prediction but also provides a novel geometric deep learning framework for the general non-Euclidean data analysis. 
\end{abstract}



\section{Introduction}\label{sec1}

Drug development is a complex and costly process, typically requiring decades of time and substantial investment \cite{Hughes2011}. In this challenging landscape, artificial intelligence (AI) has become particularly valuable, significantly impacting the prediction of molecular properties and showing immense promise in drug design \cite{zhang2024attention, chan2019advancing}. AI has greatly advanced virtual screening processes, potentially reducing the time and investment required \cite{carpenter2018machine, maia2020structure, zhao2021identifying}. AI-based molecular models, which drive these advancements, generally fall into two categories: those based on molecular descriptors and end-to-end deep learning models \cite{xia2024understanding}. 

The first category relies on molecular descriptors or fingerprints as input features for machine learning algorithms. This process, known as featurization or feature engineering, involves not only capturing physical, chemical, and biological properties but also incorporating a wide array of fingerprints based on molecular structure information. Among these, structure-based fingerprints, particularly those derived from topological data analysis methods, have proven highly effective in molecular representation and featurization \cite{Duran2018, nguyen2020review, Bonner2022, meng2021persistent}. The integration of these fingerprints with learning models has achieved significant success in various stages of drug design, including protein-ligand binding affinity prediction \cite{nguyen2019agl, Szulc2023}, protein mutation analysis \cite{chen2020mutations, wang2020mutations}, and drug design \cite{gao2020generative,zhao2022geometric}, among other areas.

The second category includes end-to-end deep learning models that utilize various molecular representations such as Simplified Molecular Input Line Entry System (SMILES) strings, images, or molecular graphs, and deploy architectures like Transformers, Convolutional Neural Networks (CNNs), and Graph Neural Networks (GNNs) for prediction \cite{li2021trimnet,li2022multiphysical, Kang2022, cai2022fp, liu2022attention,zhang2024mvmrl}. Among these, molecular graphs based on covalent bonds are predominantly employed to describe molecular topology. Geometric deep learning models based on these molecular graphs, such as Graph Convolutional Networks (GCNs) \cite{mercado2021graph}, graph autoencoders \cite{liu2018constrained}, and graph transformers \cite{rong2020self}, have been extensively used in molecular data analysis and drug design. Additionally, recent research has demonstrated that molecular descriptors based on non-covalent interactions perform exceptionally well in predicting protein-ligand and protein-protein binding affinities \cite{wang2020topology}. These observations imply that new geometric molecular graph representations could surpass traditional covalent-bond-based graphs. By integrating these geometry-based molecular graphs into Geometric Deep Learning (GDL) models, it is possible to enhance model performance and deepen the understanding of molecular geometry \cite{shen2023molecular}.

\begin{figure*}[htbp]
    \centering
    \includegraphics[width=1.0\textwidth]{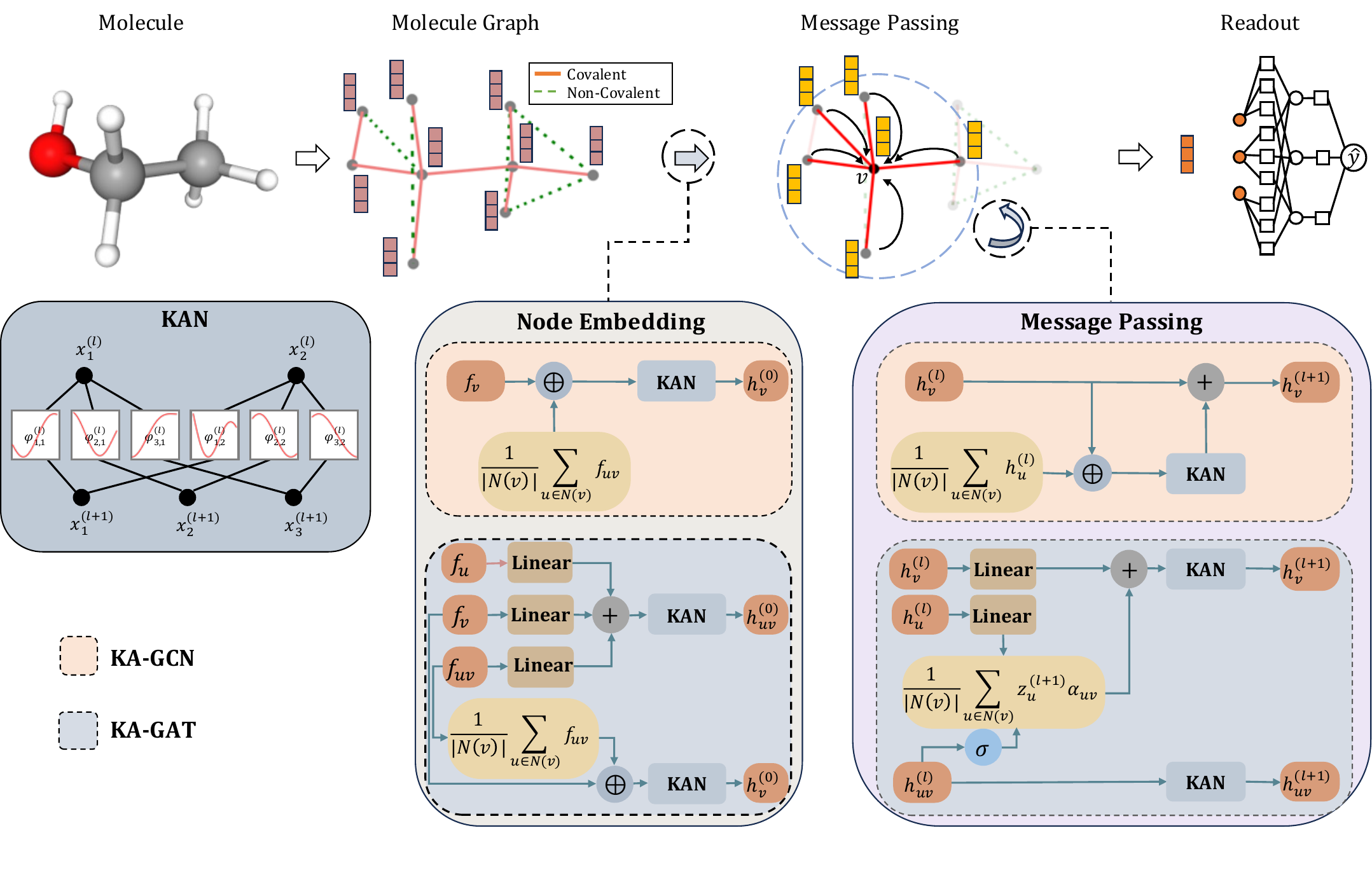} 
    \caption{Overview of the KA-GNN model architecture. The flowchart illustrates the modified components within the GNN: node embedding, message-passing, pooling and prediction modules.}
    \label{fig:KA-GNN}
\end{figure*}

Kolmogorov-Arnold Networks (KANs), which are based on the Kolmogorov-Arnold representation theorem, are increasingly recognized as a potent alternative to Multi-layer Perceptrons (MLPs). KANs distinguished by their unique architecture that employs different learnable activation functions, eliminate traditional linear weight matrices and enhance model accuracy and efficiency, particularly in solving partial differential equations, as described by Liu et al. \cite{liu2024kan}. Recent research highlights the versatility and adaptability of KANs across various domains \cite{Bresson2024Kagnns, Koenig2024KANODEs, Liu2024ikan, genet2024tkan}. One notable application is the integration of KANs with existing neural network models to enhance performance. For instance, Genet et al. \cite{genet2024tkan} significantly improved multi-step time series forecasting by integrating KANs with Long Short-Term Memory networks (LSTMs). Cheon et al. \cite{cheon2024kolmogorov} effectively classified remote sensing scenes by merging KANs with pre-trained Convolutional Neural Network (CNN) models. Kiamari et al. \cite{kiamari2024gkan} demonstrated their multifunctionality by incorporating KANs into Graph Convolutional Networks (GCNs) for semi-supervised graph learning tasks. Furthermore, adaptations in the base functions of KANs to better suit neural network applications have led to significant enhancements. Bozorgasl et al. \cite{bozorgasl2024wav} improved the interpretability and performance of KANs by incorporating wavelet functions that more effectively capture the frequency components of data. Tashin et al. \cite{ahmed2024graphkan} employed KANs with Fourier transform basis functions for feature transformation before GNN processing, validating their utility in Small Molecule-Protein Interaction Predictions. Li et al. \cite{li2024} used adaptive Radial Basis Functions (RBFs) in KANs to enhance feature updating and replace MLPs in the prediction phase, demonstrating robust integration capabilities with various neural network frameworks. Kashefi et al. \cite{kashefi2024} applied Jacobi polynomials to design KAN layers for GNNs, effectively predicting fluid fields on irregular geometries. These advancements underscore the significant potential of KANs to refine neural network architectures. The ongoing exploration of how to further enhance KANs and integrate it into node feature embedding, message passing in different GNN frameworks, and the prediction phase remains a vital area of research.

In this paper, we introduce the first non-trivial Kolmogorov-Arnold Network-based Graph Neural Networks (KA-GNNs), including KAN-based convolutional networks (KA-GCN) and KAN-based graph attention network (KA-GAT). Figure \ref{fig:KA-GNN} outlines the general KA-GNNs architecture. Different from all the previous trivial KAN-based GNN models, which only replace the MLP in the readout part with a standard KAN module, we utilizes KAN to optimize GNN architectures at three major levels, including node embedding, message passing, and readout. Further, a Fourier series-based KAN model has been developed and a rigorous theoretical prove of its robust approximation capability is given. Based on seven benchmark datasets, we have extensively validated our KA-GNNs and compared with state-of-the-art models. Our KA-GNNs have achieved both great accuracy and efficiency, providing a novel geometric deep learning framework for analyzing general non-Euclidean data. 

\section{Results}\label{sec2}

\subsection{Kolmogorov-Arnold Network (KAN)}
\textbf{Kolmogorov-Arnold Representation Theorem}
The Kolmogorov–Arnold Representation Theorem (or Superposition Theorem) is a milestone in the field of real analysis and approximation theory. It states that every multivariate continuous function can be represented as superposition of the addition of continuous functions of one variable. This theorem not only solves Hilbert's thirteenth problem itself but also generalizes it to a broader form.

Vladimir Arnold and Andrey Kolmogorov's works \cite{kolmogorov1957representation} prove that arbitrary multivariate continuous function \( f \) can be written as a finite composition of continuous functions of a single variable and the binary operation of addition. More specifically,
\begin{equation}
f(x_1, \ldots, x_n) = \sum_{q=0}^{2n+1} \Phi_q \left( \sum_{p=1}^{n} \phi_{q,p}(x_p) \right),
\label{KA}
\end{equation}
where \( n \) denote the number of variables of the function \( f \). \( \Phi_q : \mathbb{R} \to \mathbb{R} \) and \( \phi_{q,p} : [0,1] \to \mathbb{R} \) are continues function.

\subsubsection{Kolmogorov-Arnold Network (KAN)}
Inspired by the Kolmogorov-Arnold representation theorem, Liu et al. \cite{liu2024kan} proposed a new deep learning architecture called the Kolmogorov-Arnold Network (KAN) as a promising alternative to the Multi-Layer Perceptron (MLP). To enhance the KAN's representational capacity and leverage modern techniques (e.g., backpropagation) for training the networks, KAN transcends several limitations of the Kolmogorov-Arnold representation theorem:

\begin{itemize}
    \item KAN does not adhere to the original depth-2 width-$(2n + 1)$ representation; instead, it generalizes to arbitrary widths and depths. Specifically, let the activation values in layer $l$ be denoted by $\mathbf{x}^{(l)} := (x_{1}^{(l)}, x_{2}^{(l)}, \cdots, x_{n_l}^{(l)})$, where $n_l$ is the width of layer $l$. The activation value in layer $l+1$ is then simply the sum of all incoming post-activations:
    \begin{equation}
        x_{j}^{(l+1)} = \sum_{i=1}^{n_l} \phi_{j,i}^{(l)}(x_{i}^{(l)}), \quad j = 1, \ldots, n_{l+1}. 
    \label{kan:update}
    \end{equation}
    Here, $\phi_{j,i}^{(l)}$ for $i=1, \ldots, n_l$ and $j = 1, \ldots, n_{l+1}$ are the pre-activation functions in layer $l$. The roles of these functions in KAN are analogous to the roles of the inner functions $\phi_{q,p}$ in equation \ref{KA}.
    \item Although many constructive proofs of the Kolmogorov-Arnold representation theorem indicate that the inner functions $\phi_{q,p}$ in equation \ref{KA} are highly non-smooth \cite{sprecher1996numerical,braun2009constructive}, KAN opts for smooth functions as pre-activation functions $\phi_{j,i}^{(l)}$ to facilitate backpropagation. Liu et al. \cite{liu2024kan} selected functions based on B-splines. 
\end{itemize}

\subsubsection{Fourier-series KAN model}
To optimize the network and avoid complex calculations, we propose to utilize Fourier series as the pre-activation functions for KAN \cite{ahmed2024graphkan} as in Eq.(\ref{kan:update}):
\begin{equation}
\phi_{j,i}^{(l)}(x) =\sum_{k=1}^K\left( A_{k,j,i}^{(l)}\cos(kx) + B_{k,j,i}^{(l)}\sin(kx)\right),
\label{pre:activation}
\end{equation}
where \(K\) is the number of harmonics, and \(A_{k,j,i}^{(l)}\) and \(B_{k,j,i}^{(l)}\) are learnable parameters initially sampled from a normal distribution \(\text{N}\left(0, \frac{1}{n_{l+1} \times K}\right)\).

Consequently, the activation value at the $j$-th neuron in layer \(l + 1\) can be obtained by,
\begin{equation}
x_j^{(l+1)} = \sum_{i=1}^{n_l}\sum_{k=1}^K \left(A_{k,j,i}^{(l)}\cos(kx_i^{(l)}) + B_{k,j,i}^{(l)}\sin(kx_i^{(l)})\right),
\label{kan:arc}
\end{equation}
where \(\mathbf{x}^{(l)} = (x_1^{(l)}, x_2^{(l)}, \dots, x_{n_l}^{(l)})\) denotes the input vector of activation values in layer \(l\) of the KAN, and \(\mathbf{x}^{(l+1)} = (x_1^{(l+1)}, x_2^{(l+1)}, \dots, x_{n_{l+1}}^{(l+1)})\) represents the output vector. Eq (\ref{kan:arc}) can be concisely expressed as: \(\mathbf{x}^{(l+1)} = \text{KAN}_l(\mathbf{x}^{(l)})\), where \(\text{KAN}_l(\cdot)\) denotes the above KAN function in layer \(l\). This network is integrated into a graph neural network, culminating in a novel architecture named KA-GNNs, which is employed for molecular property prediction tasks.

Given that the inner functions in Eq (\ref{KA}) of the Kolmogorov-Arnold representation theorem can exhibit significant non-smoothness, we do not use this theorem as the foundational theory for showing the approximation capability of our model. Instead, we base our approach on the extension of Carleson's theorem \cite{carleson1966convergence} regarding the convergence of Fourier series for multivariable functions, as proved by Fefferman in \cite{fefferman1971convergence}:

\begin{theorem}
Let \( \mathbf{Z}^n \) denote the \( n \)-dimensional integer lattice, and let \( Z_N^n = \{1, 2, \ldots, N\}^n \subset \mathbf{Z}^n \). Then for the function \( f \in L^2([0, 2\pi]^n) \) and its Fourier expansion:
\[
f(\vec{\mathbf{x}}) \sim \sum_{\vec{\mathbf{k}} \in \mathbf{Z}^n } \left( a_{\vec{\mathbf{k}}} \cos(\vec{\mathbf{k}} \cdot \vec{\mathbf{x}}) + b_{\vec{\mathbf{k}}} \sin(\vec{\mathbf{k}} \cdot \vec{\mathbf{x}}) \right),
\]
where \(x\in [0, 2\pi]^n \text{ and }  L^2([0, 2\pi]^n)\) denotes the space of square-integrable functions on \( [0, 2\pi]^n \), which consists of all functions \( f \) such that \( \int_{[0, 2\pi]^n} |f(\vec{\mathbf{x}})|^2 \, d\vec{\mathbf{x}} < \infty \). We have
\[
f(\vec{\mathbf{x}}) = \lim_{N \to \infty} \sum_{\vec{\mathbf{k}} \in Z_N^n} \left( a_{\vec{\mathbf{k}}} \cos(\vec{\mathbf{k}} \cdot \vec{\mathbf{x}}) + b_{\vec{\mathbf{k}}} \sin(\vec{\mathbf{k}} \cdot \vec{\mathbf{x}}) \right)
\]
almost everywhere.
\end{theorem}

The above theorem demonstrates the strong approximation capability of Fourier series, which is why we adopt Fourier series as the foundational basis for our model. Therefore, we retain the architecture of the Kolmogorov-Arnold network but replace the pre-activation functions with Fourier series. We can further prove that this new KAN architecture provides the potential of robust approximation capability:

\begin{theorem}
Let \( f \in L^2([0, 2\pi]^n) \) be a square-integrable function. For almost every \(\vec{x} \in [0, 2\pi]^n\) and for any \(\epsilon > 0\), there exist a positive integer \(K\) and a sequence of Fourier-based KAN functions \( \mathrm{KAN}_l \) at multiple layers \( l = 0, 1, \ldots, L \), such that the number of harmonics in the pre-activation functions of these KAN functions does not exceed \(K\), and
\[
\left| f(\vec{x}) - \mathrm{KAN}_L \circ \mathrm{KAN}_{L-1} \circ \cdots \circ \mathrm{KAN}_0(\vec{x}) \right| < \epsilon,
\]
where \(\circ\) denotes function composition.
\end{theorem}

In summary, our model extends the Kolmogorov-Arnold Network (KAN) architecture by incorporating Fourier series as pre-activation functions to enhance approximation capability. 

\subsection{Kolmogorov-Arnold Network-based Graph Neural Network Models}

\begin{table*}[htbp]
  \centering

  \caption{Features Type and Description in KA-GNN models. Here we list both atom, covalent bond, and non-covalent bond features.}
  \label{tab:feature}
  \resizebox{\textwidth}{!}{  
  \begin{tabular}{lcccc} 
    \toprule
    \multicolumn{2}{c}{\textbf{Features Type}} & \textbf{Description} & \textbf{Type} & \textbf{Size} \\
\midrule
    \multirow{1}{*}{\begin{tabular}[c]{@{}c@{}}Atom\end{tabular}}
& CGCNN&  Atomic number, Radius, and electronegativity & One-Hot & 92 \\
\midrule
\multirow{4}{*}{\begin{tabular}[c]{@{}c@{}}Covalent Bond\\ \end{tabular}}
& Bond Directionality& None, Beginwedge, Begindash, etc. & One-Hot & 7 \\
&Bond Type & Single, Double, Triple, or Aromatic. & One-Hot & 4 \\
&Bond Length & Numerical and square length of the bond. & Float & 2 \\
&In Ring & Indicates if the bond is part of a chemical ring. & One-Hot & 2 \\
\midrule
\multirow{2}{*}{\begin{tabular}[c]{@{}c@{}}Non-Covalent Bond\\ \end{tabular}}
& Atom charges& Atoms charges in Molecular ($q_i$, $q_j$, $q_i$$\cdot$ $q_j$) & Float & 3 \\
&Distance between atoms & Distance between atoms ($\frac{1}{d_{ij}}$, $\frac{1}{d_{ij}^6}$, $\frac{1}{d_{ij}^{12}}$) & Float & 3 \\ 
\bottomrule
  \end{tabular}}
\end{table*}

In our KA-GNN models, a molecule is represented as a graph \(G = (V, E)\), where \(V\) denotes the set of nodes and \(E\) denotes the set of edges. Each node represents an atom and an edge is formed among any two atoms if their distance is within a cutoff distance (in our model, we use cutoff distance as 5 \AA). Figure \ref{fig:KA-GNN} \textbf{A} illustrates the complex interactions within the molecule, highlighting both the covalent bonds (solid lines) and non-covalent cut-off bonds (dashed lines) with distances less than 5 angstroms are considered in our KA-GNN model. Our atomic features—comprising atomic number, radius, and electronegativity—are derived using Rdkit, following the approach in CGCNN \cite{xie2018crystal} and PCNN \cite{li2024pathcomplex}. Each node \(v \in V\) is associated with a feature vector \(f_v\), which is a 92-dimensional vector composed of one-hot encoded representations of atomic properties, following the approach described in \cite{xie2018crystal}. Similarly, each edge \(uv \in E\) is associated with a feature vector \(f_{uv}\), which is a 21-dimensional vector incorporating both chemical and geometrical information of the bond \(uv\). The edges in our model can be classified into two types, i.e., covalent bonds and non-covalent bonds, with different initial features. Detailed descriptions of the feature vectors are provided in Table \ref{tab:feature}.

\subsubsection{KA-GCN Model}
Our KA-GCN architecture has three major steps, including node embedding, message-passing, and readout. First, one KAN layer is employed in the initialization of node embeddings \(h_v^{(0)}, v \in V\) as follows,
\begin{equation}
    h_v^{(0)} := \text{KAN}\left(f_v\oplus (\frac{1}{|N(v)|} \sum_{u \in N(v)} f_{uv})\right),
\end{equation}
where \(N(v)\) represents the set of the neighboring nodes of node \(v\) (excluding \(v\) itself), \(|N(v)|\) is the number of neighboring nodes, and \( \oplus \) represents vector concatenation.

Second, one or several KAN layers are employed in the message passing module. At the layer \(l\), the KAN-based message passing module can be expressed as,
\begin{equation}
    h_v^{(l+1)} := h_v^{(l)} + \text{KAN}\left(h_v^{(l)} \oplus( \frac{1}{|N(v)|}\sum_{u \in N(v)} h_u^{(l)})\right).
\end{equation}
Computationally, the first KAN layer maps initial feature vector \(f_v\) from dimension 113 to updated feature vector \(h_v\) with dimension 64. After that, the feature dimension will stay the same in the later message-passing process.

Third, one or several KAN layers are employed in the readout module. After $L$-th the message passing phases, we apply average pooling across all node features to obtain a latent vector for the molecular graph \(G\). An one-layer or two-layer KAN model is employed in readout module for the final prediction as follows,
\begin{equation*}
    \hat{y} := \text{KAN}\left( \frac{1}{|V|}\sum_{v\in V} h_v^{(L)} \right)\text{or}, \hat{y} := \text{KAN}\left(\text{KAN}\left( \frac{1}{|V|}\sum_{v\in V} h_v^{(L)} \right)\right)  
\end{equation*}

The cross-entropy is used as the loss function as follows,
\begin{equation*}
    \mathcal{L} := -\sum_{i} \left( y_i \log(\hat{y}_i) + (1 - y_i) \log(1 - \hat{y}_i) \right),
\end{equation*}
where \(y_i\) denotes the actual label, and \(\hat{y}_i\) denotes the predicted label. 


\subsubsection{KA-GAT Model} 
Similar to the KA-GCN architecture, the KA-GAT model initializes the node and edge embeddings, \( h_v^{(0)} \) and \( h_{uv}^{(0)} \), using a KAN layer as follows:
\begin{equation}
\begin{split}
    h_v^{(0)} &:= \text{KAN}\left(f_v \oplus \left(\frac{1}{|N(v)|} \sum_{u \in N(v)} f_{uv}\right)\right),\\
    h_{uv}^{(0)} &:= \text{KAN}\left(\text{Linear}(f_v) + \text{Linear}(f_{uv}) + \text{Linear}(f_u)\right)
\end{split}
\end{equation}

Subsequently, the KA-GAT model uses these initial embeddings to compute attention scores that guide the aggregation of messages:

\begin{equation}
    \begin{split}
        z_v^{(l+1)} &:= \text{Linear}\left(h_v^{(l)}\right), z_u^{(l+1)} :=\text{Linear}\left(h_u^{(l)}\right),\\
        \alpha_{uv} &:= \text{Softmax}\left(h_{uv}^{(l)}\right), m_{uv}^{(l+1)} :=  z_u^{(l+1)}  \cdot \alpha_{uv}\\
        m_v^{(l+1)} &:= z_v^{(l+1)} + \frac{1}{|N(v)|}\left(\sum_{u \in N(v)} m_{uv}^{(l+1)} \right)
    \end{split}
\end{equation}
Then, one or more KAN layers are used to update the node and edge features:
\begin{equation}
    h_v^{(l+1)} := \text{KAN}(m_v^{(l+1)}), \quad h_{uv}^{(l+1)} := \text{KAN}(h_{uv}^{(l)})
\end{equation}

Finally, after message passing, the KA-GAT model employs the same readout module as used in the KA-GCN.

\subsection{KA-GNNs for Molecular Property Analysis} 

\subsubsection{Performance and comparison of KA-GNNs}
To evaluate the performance of our proposed KA-GNN models, we consider seven widely-used benchmark datasets from MoleculeNet \cite{wu2018moleculenet}. Of these, three datasets—MUV, HIV, and BACE—are from the biophysics domain. The rest four datasets are from the physiology domain, including BBBP, Tox21, SIDER, and ClinTox. 

In our comparative analysis across seven datasets, we evaluated a range of geometric deep learning models, including AttentiveFP \cite{xiong2019pushing}, D-MPNN \cite{yang2019analyzing}, Mol-GDL \cite{shen2023molecular}, N-Gram \cite{liu2019n}, PretrainGNN \cite{hu2019strategies}, GROVER \cite{rong2020self}, GraphMVP \cite{liu2022pretraining}, MolCLR \cite{wang2022molecular}, GEM \cite{fang2022geometry}, Uni-mol \cite{zhou2023unimol}, SMPT \cite{LI2024108023}, and GNN-SKAN \cite{li2024}. Additionally, we included our KAN-based GNN models, such as the GNN-SKAN \cite{li2024}. Further details on these models can be found in Subsection \hyperref[sec:baseline]{Baseline Models}.

The comparative results in Table \ref{tab:mol_kan_res} further confirm the superiority of the KA-GNNs model. Our model achieves state-of-the-art performance across all benchmark datasets, particularly excelling on complex and challenging datasets such as ClinTox and MUV. These results demonstrate the robust capability of our model in handling molecular data. Notably, in the BBBP dataset, the AUC for KA-GCN and KA-GAT showed improvements of approximately 7.95\% and 7.68\%.

\begin{table*}[htbp]
\centering
\caption{Comparison of KA-GNNs with state-of-the-art geometric deep learning models in different molecular property datasets. The performance metric is the average of Receiver Operating Characteristic - Area Under the Curve (ROC-AUC). The standard deviation results are denoted as subscripts. The best model for each category is highlighted in bold, while the second-best performance is marked with underline. The notation \text{-} indicates that the results were not reported.}
\label{tab:mol_kan_res}
\resizebox{\textwidth}{!}{  
\begin{tabular}{@{}lcccccccc@{}}
\toprule
Model & BACE & BBBP & ClinTox & SIDER & Tox21 & HIV & MUV \\
No. molecules & 1513 & 2039 & 1478 & 1427 & 7831 & 41127 & 93808 \\
No. avg atoms & 65 & 46 & 50.58 & 65 & 36 & 46 & 43 \\
No. tasks & 1 & 1 & 2 & 27 & 12 & 1 & 17 \\
\midrule
D-MPNN \cite{yang2019analyzing} & 0.809$_{(0.006)}$ & 0.710$_{(0.003)}$ & 0.906$_{(0.007)}$ & 0.570$_{(0.007)}$ & 0.759$_{(0.007)}$ & 0.771$_{(0.005)}$ & 0.786$_{(0.014)}$ \\
AttentiveFP \cite{xiong2019pushing} & 0.784$_{(0.022)}$ & 0.663$_{(0.018)}$ & 0.847$_{(0.003)}$ & 0.606$_{(0.032)}$ & 0.781$_{(0.005)}$ & 0.757$_{(0.014)}$ & 0.786$_{(0.015)}$ \\
N-GramRF \cite{liu2019n} & 0.779$_{(0.015)}$ & 0.697$_{(0.006)}$ & 0.775$_{(0.040)}$ & 0.668$_{(0.007)}$ & 0.743$_{(0.009)}$ & 0.772$_{(0.004)}$ & 0.769$_{(0.002)}$ \\
N-GramXGB \cite{liu2019n} & 0.791$_{(0.013)}$ & 0.691$_{(0.008)}$ & 0.875$_{(0.027)}$ & 0.655$_{(0.007)}$ & 0.758$_{(0.009)}$ & 0.787$_{(0.004)}$ & 0.748$_{(0.002)}$ \\
PretrainGNN \cite{hu2019strategies}& 0.845$_{(0.007)}$ & 0.687$_{(0.013)}$ & 0.726$_{(0.015)}$ & 0.627$_{(0.008)}$ & 0.781$_{(0.006)}$ & 0.799$_{(0.007)}$ & 0.813$_{(0.021)}$ \\
GROVE\_base \cite{rong2020self} & 0.821$_{(0.007)}$ & 0.700$_{(0.001)}$ & 0.812$_{(0.030)}$ & 0.648$_{(0.006)}$& 0.743$_{(0.001)}$ & 0.625$_{(0.009)}$ & 0.673$_{(0.018)}$ \\
GROVE\_large \cite{rong2020self} & 0.810$_{(0.014)}$ & 0.695$_{(0.001)}$ & 0.762$_{(0.037)}$ & 0.654$_{(0.001)}$ & 0.735$_{(0.001)}$ & 0.682$_{(0.011)}$ & 0.673$_{(0.018)}$ \\
GraphMVP \cite{liu2022pretraining}& 0.812 $_{(0.009)}$ & 0.724 $_{(0.016)}$ & 0.791 $_{(0.028)}$ & 0.639 $_{(0.012)}$ & 0.759 $_{(0.005)}$ & 0.770 $_{(0.012)}$ & 0.777 $_{(0.006)}$  \\
MolCLR \cite{wang2022molecular}& 0.824 $_{(0.009)}$ & 0.722 $_{(0.021)}$ & 0.912 $_{(0.035)}$ & 0.589 $_{(0.014)}$ & 0.750 $_{(0.002)}$ & 0.781 $_{(0.005)}$ & 0.796 $_{(0.019)}$ \\
GEM \cite{fang2022geometry}& 0.856$_{(0.011)}$ & 0.724$_{(0.004)}$ & 0.901$_{(0.013)}$ & 0.672$_{(0.004)}$ & 0.781$_{(0.001)}$ & 0.806$_{(0.009)}$ & 0.817$_{(0.005)}$ \\
Mol-GDL \cite{shen2023molecular} & 0.863$_{(0.019)}$ & 0.728$_{(0.019)}$ & 0.966$_{(0.002)}$ & 0.831$_{(0.002)}$ &0.794$_{(0.005)}$ & 0.808$_{(0.007)}$ & 0.675$_{(0.014)}$ \\
Uni-mol\cite{zhou2023unimol} & 0.857 $_{(0.002)}$ & 0.729 $_{(0.006)}$ & 0.919 $_{(0.018)}$ & 0.659 $_{(0.013)}$ & 0.796 $_{(0.005)}$ & 0.808 $_{(0.003)}$ & 0.821 $_{(0.013)}$ \\
SMPT\cite{LI2024108023}& 0.873 $_{(0.015)}$ & 0.734 $_{(0.003)}$ & 0.927 $_{(0.002)}$ & 0.676 $_{(0.050)}$ & 0.797 $_{(0.001)}$ & 0.812 $_{(0.001)}$ & 0.822 $_{(0.008)}$  \\ 
\midrule
GNN-SKAN \cite{li2024}&0.747$_{(0.009)}$& 0.676$_{(0.014)}$&-&0.614$_{(0.005)}$&0.747$_{(0.005)}$&0.786$_{(0.015)}$&-\\
\textbf{KA-GCN}  & \textbf{0.890}$_{(0.014)}$ & \textbf{0.787}$_{(0.014)} $& \underline{0.989}$_{(0.003)}$ & \underline{0.842}$_{(0.001)}$ & \underline{0.799}$_{(0.005)}$ & \underline{0.821}$_{(0.005)} $& \textbf{0.834}$_{(0.009)}$ \\
\textbf{KA-GAT} &\underline{0.884}$_{(0.004)}$ & \underline{0.785}$_{(0.021)} $& \textbf{0.991}$_{(0.005)}$ &\textbf{0.847}$_{(0.002)}$& \textbf{0.800}$_{(0.006)}$ &\textbf{0.823}$_{(0.002)}$ &\underline{0.834}$_{(0.010)}$ \\
\bottomrule
\end{tabular}}
\end{table*}

\begin{table*}[htbp]
\centering
\caption{Comparison of the performance of KA-GNN (and KA-GAT) models with three types of base functions, including B-spline, polynomial, and Fourier series.}
\label{tab:MLP_GAT}
\resizebox{\textwidth}{!}{  
\begin{tabular}{lcccccc}
\toprule
\textbf{Dataset} & \multicolumn{3}{c}{\textbf{KA-GCN Models }} & \multicolumn{3}{c}{\textbf{KA-GAT Models }} \\
\cmidrule(lr){2-4} \cmidrule(lr){5-7}
 & \textbf{B-spline} & \textbf{Polynomial } & \textbf{Fourier series}  & \textbf{B-spline} & \textbf{Polynomial} & \textbf{Fourier series} \\
\midrule
BACE   & 0.771$_{(0.012)}$ & 0.853$_{(0.027)}$ & \textbf{0.890}$_{(0.014)}$  & 0.808$_{(0.009)}$ & 0.8319$_{(0.007)}$ & \textbf{0.884}$_{(0.004)}$ \\
BBBP     & 0.723$_{(0.008)}$ & 0.654$_{(0.009)}$ & \textbf{0.787}$_{(0.014)}$ & 0.657$_{(0.004)}$ & 0.708$_{(0.005)}$ & \textbf{0.785}$_{(0.021)}$ \\
ChinTox  & 0.973$_{(0.006)}$ & 0.981$_{(0.004)}$ & \textbf{0.989}$_{(0.003)}$  & 0.948$_{(0.003)}$ & 0.983$_{(0.003)}$ & \textbf{0.991}$_{(0.005)}$ \\
SIDER   & 0.824$_{(0.003)}$ & 0.832$_{(0.006)}$ & \textbf{0.842}$_{(0.001)}$  & 0.825$_{(0.004)}$ & 0.836$_{(0.001)}$ & \textbf{0.847}$_{(0.002)}$ \\
Tox21   & 0.724$_{(0.005)}$ & 0.715$_{(0.004)}$ & \textbf{0.799}$_{(0.005)}$  & 0.731$_{(0.012)}$ & 0.753$_{(0.004)}$ & \textbf{0.800}$_{(0.006)}$ \\
HIV    & 0.753$_{(0.007)}$ & 0.804$_{(0.010)}$ & \textbf{0.821}$_{(0.005)}$ & 0.744$_{(0.018)}$ & 0.818$_{(0.006)}$ & \textbf{0.823}$_{(0.002)}$ \\
MUV     & 0.638$_{(0.008)}$ & 0.787$_{(0.012)}$ & \textbf{0.834}$_{(0.009)}$  & 0.792$_{(0.013)}$ & 0.797$_{(0.011)}$ & \textbf{0.834}$_{(0.010)}$ \\
\bottomrule
\end{tabular}}
\end{table*}

\subsubsection{Importance of Fourier series for KA-GNNs}
In this research, we employed a Fourier-based Kolmogorov-Arnold Network (KAN) as the primary computational mechanism within our GNN architecture, which differs significantly from traditional MLP-based approaches. To assess the impact of different base functions on the prediction of molecular properties, we explored several base function including B-spline, polynomial, Fourier-series, in the KA-GCNs architecture. These models were tested within the traditional GCN framework as well as the GAT framework. The comparative results are summarized in Table \ref{tab:MLP_GAT}. The Polynomial-base function can be expressed in Eq (\ref{kan:pol}).

\begin{equation}
x_j^{(l+1)} = \sum_{i=1}^{n_l}\sum_{k=1}^K \left(C_{k,j,i}^{(l)}(x_i^{(l)})^k\right),
\label{kan:pol}
\end{equation}
where, \(C_{k,j,i}^{(l)}\) is learnable parameters initially sampled from a normal distribution \(\mathcal{N}\left(0, \frac{1}{n_{l+1} \times K}\right)\).  \(\mathbf{x}^{(l)} = (x_1^{(l)}, x_2^{(l)}, \dots, x_{n_l}^{(l)})\) denotes the input vector of activation values in layer \(l\) of the Polynomial-series based KAN.

\begin{figure*}[htbp]
    \centering
    \includegraphics[width=\linewidth]{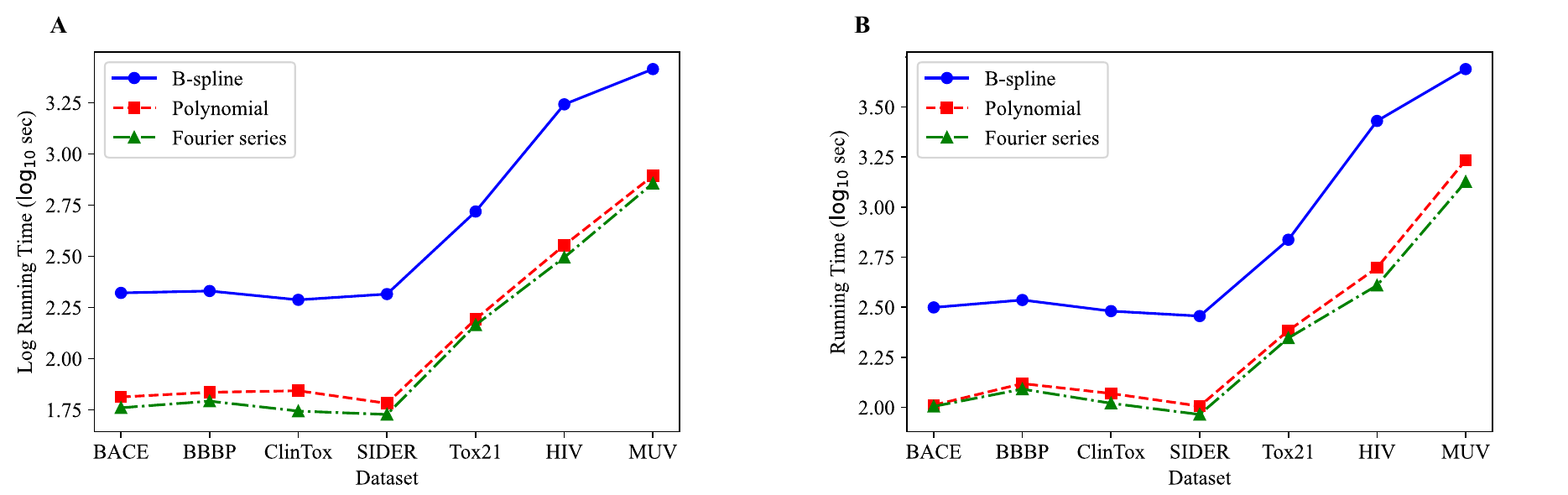}
    \caption{The comparison the computational efficiency of KA-GNNs based on B-spline, polynomial, and Fourier series. A. Running time of KA-GCN model for 100 epochs across different datasets. B. Running time of KA-GAT Model for 100 epochs across different datasets.}
    \label{fig:efficiency}
\end{figure*}

The data in the table illustrate that the KA-GCN and KA-GAT models, which incorporate Fourier-based methods, consistently outperform other models across all tested datasets. This underscores the substantial advantages of integrating Fourier-based approaches in enhancing the predictive accuracy and generalizability of GNN architectures. The superiority of these models is largely attributable to the powerful approximation capabilities of Fourier-based KAN, which we have substantiated in Theorem 2 and will further validate through subsequent function fitting evaluations. Furthermore, our ablation study not only validates our hypothesis that Fourier-based KAN provides a robust framework for improving molecular property prediction, significantly enhancing everything from molecular feature initialization to message passing in graph neural networks and prediction computations, but also compares the operational efficiency of models under different base functions. As shown in Figure \ref{fig:efficiency} A and B, the results demonstrate that Fourier-based KAN significantly outperforms B-spline KAN in terms of efficiency. The Fourier-based models also clearly surpass the GCN and GAT models that utilize these other two base functions.

\begin{table*}[htbp]
\centering
\caption{Comparison of GCN/GAT models and KA-GCN/KA-GAT with Fourier series.}
\label{tab:MLP_Fourier}
\begin{tabular}{lcccc}
\toprule
\textbf{Dataset} & \textbf{GCN} & \textbf{KA-GCN} & \textbf{GAT} & \textbf{KA-GAT} \\
\midrule
BACE    & 0.835$_{(0.014)}$ & \textbf{0.890}$_{(0.014)}$ & 0.834$_{(0.012)}$ & \textbf{0.884}$_{(0.004)}$ \\
BBBP    & 0.735$_{(0.011)}$ & \textbf{0.787}$_{(0.014)}$ & 0.707$_{(0.007)}$ & \textbf{0.785}$_{(0.021)}$ \\
ChinTox & 0.979$_{(0.004)}$ & \textbf{0.989}$_{(0.003)}$ & 0.983$_{(0.006)}$ & \textbf{0.991}$_{(0.005)}$ \\
SIDER   & 0.834$_{(0.001)}$ & \textbf{0.842}$_{(0.001)}$ & 0.836$_{(0.002)}$ & \textbf{0.847}$_{(0.002)}$ \\
Tox21   & 0.747$_{(0.006)}$ & \textbf{0.799}$_{(0.005)}$ & 0.751$_{(0.007)}$ & \textbf{0.800}$_{(0.006)}$ \\
HIV     & 0.762$_{(0.005)}$ & \textbf{0.821}$_{(0.005)}$ & 0.761$_{(0.003)}$ & \textbf{0.823}$_{(0.002)}$ \\
MUV     & 0.741$_{(0.006)}$ & \textbf{0.834}$_{(0.009)}$ & 0.784$_{(0.019)}$ & \textbf{0.834}$_{(0.010)}$ \\
\bottomrule
\end{tabular}
\end{table*}

Furthermore, in this study, we also compared the traditional GCN and GAT models with our KA-GNNs (KA-GCN and KA-GAT), on the same feature molecular graphs. Table \ref{tab:MLP_Fourier} illustrates that KA-GNNs demonstrate significant performance improvements across all datasets. This confirms that the incorporation of Fourier series significantly enhances the models’ ability to recognize and express the inherent periodicity and complexity in chemical structures. 

\section*{Discussion}
We attribute the effectiveness and power of the KA-GNNs model to three core innovations. First, in constructing molecular graphs, we incorporate edges not only from traditional covalent bonds but also from non-covalent interactions based on a cut-off distance. This approach significantly enhances the model's understanding of molecular structures, enabling it to capture more molecular properties than traditional covalent interactions alone. Second, we integrate the Kolmogorov-Arnold Network (KAN) into our model. KAN significantly reduces the number of parameters while providing high interpretability and enhanced expressive power, leading to superior performance of the KA-GNNs model. Third, we replace the pre-activation function in the KAN model from the original B-splines to Fourier series functions. Fourier series functions, widely used in signal processing, have proven effective in our model for analyzing and learning high-dimensional graphical data. By learning and optimizing the coefficients of these Fourier series functions for different variables, KA-GNNs improves the accuracy and stability of predictions when dealing with complex chemical and biological data structures. These three innovations collectively contribute to the outstanding performance of KA-GNNs, making it a powerful tool for molecular property prediction.

However, our KA-GNNs still have limitations. One of the key challenges is their interpretability. Even though KANs have been designed for excellent interpretability and indeed show good results in PDE examples. In our study, the pruning process however does not generate biologically meaningful superposition structures, that are easy to interpretate.

\section{Methods}

\subsection{Approximation Capability of Fourier KAN}
Given that the inner functions in Eq (\ref{KA}) of the Kolmogorov-Arnold representation theorem can exhibit significant non-smoothness, we do not use this theorem as the foundational theory for showing the approximation capability of our model. Instead, we base our approach on the extension of Carleson's theorem \cite{carleson1966convergence} regarding the convergence of Fourier series for multivariable functions, as proved by Fefferman in \cite{fefferman1971convergence}:

\begin{theorem}
Let \( \mathbf{Z}^n \) denote the \( n \)-dimensional integer lattice, and let \( Z_N^n = \{1, 2, \ldots, N\}^n \subset \mathbf{Z}^n \). Then for the function \( f \in L^2([0, 2\pi]^n) \) and its Fourier expansion:

\[
f(\vec{\mathbf{x}}) \sim \sum_{\vec{\mathbf{k}} \in \mathbf{Z}^n } \left( a_{\vec{\mathbf{k}}} \cos(\vec{\mathbf{k}} \cdot \vec{\mathbf{x}}) + b_{\vec{\mathbf{k}}} \sin(\vec{\mathbf{k}} \cdot \vec{\mathbf{x}}) \right),
\]

where \(x\in [0, 2\pi]^n \text{ and }  L^2([0, 2\pi]^n)\) denotes the space of square-integrable functions on \( [0, 2\pi]^n \), which consists of all functions \( f \) such that \( \int_{[0, 2\pi]^n} |f(\vec{\mathbf{x}})|^2 \, d\vec{\mathbf{x}} < \infty \). We have

\[
f(\vec{\mathbf{x}}) = \lim_{N \to \infty} \sum_{\vec{\mathbf{k}} \in Z_N^n} \left( a_{\vec{\mathbf{k}}} \cos(\vec{\mathbf{k}} \cdot \vec{\mathbf{x}}) + b_{\vec{\mathbf{k}}} \sin(\vec{\mathbf{k}} \cdot \vec{\mathbf{x}}) \right)
\]

almost everywhere.
\end{theorem}

The above theorem demonstrates the strong approximation capability of Fourier series, which is why we adopt Fourier series as the foundational basis for our model. Therefore, we retain the architecture of the Kolmogorov-Arnold network but replace the pre-activation functions with Fourier series. We can further prove that this new KAN architecture provides the potential of robust approximation capability:

\begin{theorem}
Let \( f \in L^2([0, 2\pi]^n) \) be a square-integrable function. For almost every \(\vec{x} \in [0, 2\pi]^n\) and for any \(\epsilon > 0\), there exist a positive integer \(K\) and a sequence of Fourier-based KAN functions \( \mathrm{KAN}_l \) at multiple layers \( l = 0, 1, \ldots, L \), such that the number of harmonics in the pre-activation functions of these KAN functions does not exceed \(K\), and

\[
\left| f(\vec{x}) - \mathrm{KAN}_L \circ \mathrm{KAN}_{L-1} \circ \cdots \circ \mathrm{KAN}_0(\vec{x}) \right| < \epsilon,
\]

where \(\circ\) denotes function composition.
\end{theorem}

\begin{theorem}
Let \( \mathbf{Z}^n \) denote the \( n \)-dimensional integer lattice, and let \( Z_N^n = \{1, 2, \ldots, N\}^n \subset \mathbf{Z}^n \). Then for the function \( f \in L^2([0, 2\pi]^n) \) and its Fourier expansion:

\[
f(\vec{\mathbf{x}}) \sim \sum_{\vec{\mathbf{k}} \in \mathbf{Z}^n } \left( a_{\vec{\mathbf{k}}} \cos(\vec{\mathbf{k}} \cdot \vec{\mathbf{x}}) + b_{\vec{\mathbf{k}}} \sin(\vec{\mathbf{k}} \cdot \vec{\mathbf{x}}) \right),
\]

where \(x\in [0, 2\pi]^n \text{ and }  L^2([0, 2\pi]^n)\) denotes the space of square-integrable functions on \( [0, 2\pi]^n \), which consists of all functions \( f \) such that \( \int_{[0, 2\pi]^n} |f(\vec{\mathbf{x}})|^2 \, d\vec{\mathbf{x}} < \infty \).

We have

\[
f(\vec{\mathbf{x}}) = \lim_{N \to \infty} \sum_{\vec{\mathbf{k}} \in Z_N^n} \left( a_{\vec{\mathbf{k}}} \cos(\vec{\mathbf{k}} \cdot \vec{\mathbf{x}}) + b_{\vec{\mathbf{k}}} \sin(\vec{\mathbf{k}} \cdot \vec{\mathbf{x}}) \right)
\]

almost everywhere.
\end{theorem}

\textbf{Proof.} The theorem follows as a particular instance of the multidimensional Carleson theorem, as established in \cite{fefferman1971convergence}.

\begin{theorem}
Let \( f \in L^2([0, 2\pi]^n) \) be a square-integrable function. For almost every \(\vec{x} \in [0, 2\pi]^n\) and for any \(\epsilon > 0\), there exist a positive integer \(K\) and a sequence of Fourier-based KAN functions \( \mathrm{KAN}_l \) at multiple layers \( l = 0, 1, \ldots, L \), such that the number of harmonics in the pre-activation functions of these KAN functions does not exceed \(K\), and

\[
\left| f(\vec{x}) - \mathrm{KAN}_L \circ \mathrm{KAN}_{L-1} \circ \cdots \circ \mathrm{KAN}_0(\vec{x}) \right| < \epsilon,
\]

where \(\circ\) denotes function composition.
\end{theorem}

\textbf{Proof.} Based on \textbf{Theorem 1}, there exists a number $N$ such that

\begin{equation}
\left|f(\vec{\mathbf{x}})-\sum_{\vec{\mathbf{k}} \in Z_N^n} \left( a_{\vec{\mathbf{k}}} \cos(\vec{\mathbf{k}} \cdot \vec{\mathbf{x}}) + b_{\vec{\mathbf{k}}} \sin(\vec{\mathbf{k}} \cdot \vec{\mathbf{x}}) \right)\right| < \frac{\epsilon}{2}.
\label{eq:1}
\end{equation}

Furthermore, there exists $\delta > 0$ such that for any $\vec{\mathbf{k}} \in \mathbb{Z}_N^n$ and any $y \in \mathbb{R}$ with $\Vert y - \vec{\mathbf{k}}\cdot\vec{\mathbf{x}} \Vert < \delta$, we have 

\begin{equation}
|\cos(y) - \cos(\vec{\mathbf{k}} \cdot \vec{\mathbf{x}})| < \frac{\epsilon}{4N^n a_{\vec{\mathbf{k}}}},
\label{eq:cos}
\end{equation}
and
\begin{equation}
|\sin(y) - \sin(\vec{\mathbf{k}} \cdot \vec{\mathbf{x}})| < \frac{\epsilon}{4N^n b_{\vec{\mathbf{k}}}}.
\label{eq:sin}
\end{equation}

Let $S(N_1, id)(x)$ denote the partial sum of the first $N_1$ terms of the Fourier expansion of $f$, where $N_1$ is such that

\begin{equation}
|x - S(N_1, id)(x)| < \frac{\delta}{nN},
\label{eq:id}
\end{equation}
and $id:x\mapsto x$ is the identity function on $[0,2\pi]$.

To prove the result, select $K = \max\{N, N_1\}$. We construct KAN functions $\text{KAN}_0$ and $\text{KAN}_1$ in two layers to achieve $\left| f(\vec{\mathbf{x}}) - \text{KAN}_1 \circ \text{KAN}_0(\vec{\mathbf{x}}) \right| < \epsilon$. In the following proof, we denote the input vector at layer $l$ by $\mathbf{x}^{(l)} = (x_1^{(l)}, x_2^{(l)}, \dots)$, and let the initial input vector $\mathbf{x}^{(0)}=\vec{\mathbf{x}}$

For $Z_N^n = \{\vec{\mathbf{k}}_1, \vec{\mathbf{k}}_2, \dots, \vec{\mathbf{k}}_{N^n}\}$ with $\vec{\mathbf{k}}_j = (k_{j,1}, k_{j,2}, \dots, k_{j,n})$, define the pre-activation function at layer 0 as

\[
\phi_{j,i}^{(0)}(x) = k_{j,i} S(N_1, id)(x), \quad i = 1, 2, \dots, n, \, j = 1, 2, \dots, N^n.
\label{eq:phi0}
\]

Then the $j$-th neuron at layer 1 is $x_j^{(1)} = \sum_{i=1}^{n} k_{j,i} S(N_1, id)(x_i^{(0)})$. From \eqref{eq:id}, we have

\begin{equation}
\left|x_j^{(1)} - \vec{\mathbf{k}}_j \cdot \mathbf{x}^{(0)}\right| < \delta, \quad \text{for all } j = 1, 2, \dots, N^n.
\label{eq:e1}
\end{equation}

At layer 1, define the pre-activation function as

\[
\phi_{j,i}^{(1)}(x) = a_{\vec{\mathbf{k}}_i}\cos(x) + b_{\vec{\mathbf{k}}_i}\sin(x), \quad i = 1, 2, \dots, N^n, \, j = 1.
\label{eq:phi1}
\]

The single neuron at layer 2 is

\[
x_1^{(2)} = \sum_{i=1}^{N^n} \left( a_{\vec{\mathbf{k}}_i} \cos(x_i^{(1)}) + b_{\vec{\mathbf{k}}_i} \sin(x_i^{(1)}) \right).
\]

Using \eqref{eq:cos}, \eqref{eq:sin}, and \eqref{eq:e1}, we get

\[
\left| x_1^{(2)} - \sum_{\vec{\mathbf{k}} \in Z_N^n} \left( a_{\vec{\mathbf{k}}} \cos(\vec{\mathbf{k}} \cdot \mathbf{x}^{(0)}) + b_{\vec{\mathbf{k}}} \sin(\vec{\mathbf{k}} \cdot \mathbf{x}^{(0)}) \right) \right| < \frac{\epsilon}{2}.
\]

Finally, using the inequality \eqref{eq:1}, we conclude

\[
\left| f(\vec{\mathbf{x}}) - x_1^{(2)} \right| < \frac{\epsilon}{2} + \frac{\epsilon}{2} = \epsilon,
\]

Notice that $x_1^{(2)}=\text{KAN}_1\circ\text{KAN}_0(\mathbf{x}^{(0)})$ by definition, we complete the proof.

\begin{figure*}[htbp]
    \centering
    \includegraphics[width=1.0\textwidth]{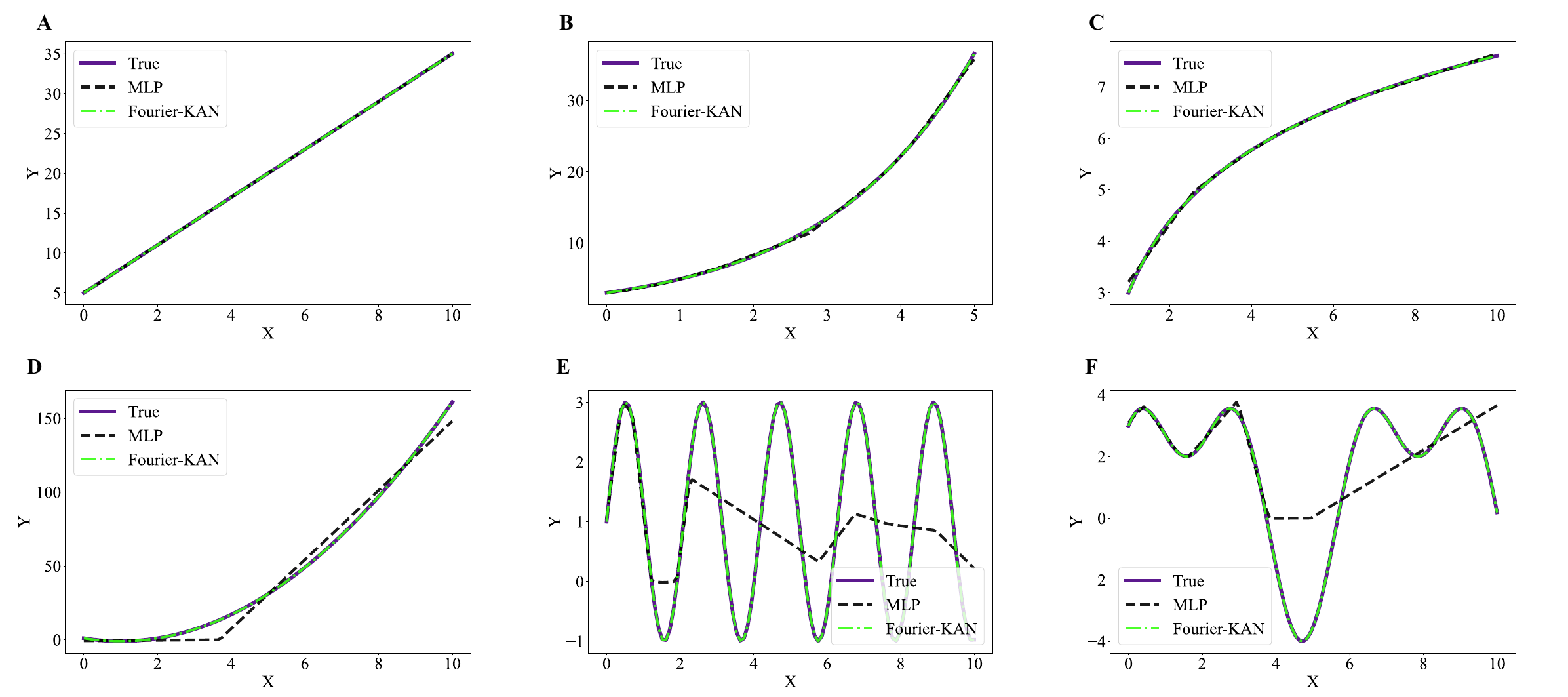} 
    \caption{Fourier-series KAN and two-layers MLP fit six different functions: A. Linear function \(y = 3x + 5\), \(K = 200\) in Fourier-series KAN; B. Exponential function \(y = 3\exp(\frac{1}{2}x)\), \(K = 120\) in Fourier-series KAN; C. Logarithmic function \(y = 2\log(x) + 3\), \(K = 100\) in Fourier-series KAN; D. Polynomial function \(y = 2x^2 - 4x + 1\), \(K = 500\) in Fourier-series KAN; E. Sin function \(y = 2\sin(3x) + 1\), \(K = 100\) in Fourier-series KAN; F. Sin and Cos function \(y = 3\sin(x) + 2\cos(2x) + 1\), \(K = 100\) in Fourier-series KAN. }
    \label{fig:KAN}
\end{figure*}

In contrast to the multilayer perceptron (MLP) approach, which increases the number of layers to enhance the model's ability to fit the data, the Fourier KAN model enhances its fitting ability by adjusting the number of Fourier basis functions (controlled by the K parameter) and the option to include a bias term to decompose complex functions into simpler sine and cosine waveforms. This strategic decomposition not only improves the model's expressiveness but also reduces the computational burden. We evaluated the fitting ability of two-layers MLP and one-layer Fourier series KAN on six different function types, and the results are shown in Figure \ref{fig:KAN}. Our analysis shows that Fourier KAN captures the underlying properties of the function more skillfully and exhibits superior expressiveness, especially in the ability to fit periodic functions.

\subsection{Benchmark Datasets}
For an extensive validation of our KA-GNN model, we used seven benchmark datasets from MoleculeNet \cite{wu2018moleculenet}. Three datasets are from the biophysics
domain: MUV, HIV, and BACE. MUV, a subset of PubChem BioAssay refined using nearest neighbor analysis, is designed for validating virtual screening techniques. The HIV dataset measures the ability of molecules to inhibit HIV replication. BACE includes both quantitative (IC50) and qualitative (binary) binding results for inhibitors of human $\beta$-secretase 1 (BACE-1). The remaining four datasets are from the physiology domain: BBBP, Tox21, SIDER, and ClinTox. BBBP contains binary labels for blood-brain barrier penetration. Tox21 provides qualitative toxicity measurements for 12 biological targets. SIDER is a database of marketed drugs and their adverse drug reactions (ADRs), grouped into 27 system organ classes. ClinTox includes qualitative data on FDA-approved drugs and those that failed clinical trials due to toxicity.

The original data in these datasets are SMILES strings of the molecules. During data preprocessing, we use the Merck molecular force field (MMFF94) function from the RDKit package to generate 3D molecular structures from the SMILES strings. Based on the generated structures, we construct graphs for the GNN in our KA-GNN model to predict the relevant properties. For evaluation, we use the Receiver Operating Characteristic - Area Under the Curve (ROC-AUC) metric. Additionally, we employ the scaffold splitting method \cite{ramsundar2019deep} to divide the datasets into training, validation, and test sets in a ratio of 8:1:1.

\begin{table}[htbp]
\centering
\caption{KA-GCN and KA-GAT parameters for different datasets.}
\small
\begin{tabular}{lccccccc}
\toprule
Dataset & Model & Batch size & LR & K & Layers & Heads&Epochs \\
\midrule
BACE & KA-GNN & 128 & 1e-4 & 1 & 3 & -& 500 \\
BACE & KA-GAT & 64 & 1e-4 & 2 & 2 & 4& 500 \\
BBBP & KA-GNN & 128 & 1e-4 & 2 & 1 & - & 500\\
BBBP & KA-GAT & 128 & 1e-4 & 5 & 2 & 2& 500\\
ChinTox & KA-GNN & 128 & 1e-4 & 2 & 2 & -& 500 \\
ChinTox & KA-GAT & 128 & 1e-4 & 5 & 2 & 2 & 500\\
SIDER & KA-GNN & 128 & 1e-4 & 2 & 1 & - & 500\\
SIDER & KA-GAT & 128 & 1e-4 & 4 & 2 & 2 & 500\\
Tox21 & KA-GNN & 512 & 1e-4 & 2 & 2 & - & 500\\
Tox21 & KA-GAT & 512 & 1e-4 & 3 & 2 & 2 & 500\\
HIV & KA-GNN & 512 & 1e-4 & 2 & 2 & - & 500\\
HIV & KA-GAT & 512 & 1e-4 & 2 & 2 & 4 & 500\\
MUV & KA-GNN & 512 & 1e-4 & 2 & 2 & - & 500\\
MUV & KA-GAT & 512 & 1e-4 & 3 & 2 & 4 & 500\\
\bottomrule
\end{tabular}
\label{tab:ka_gnn_gat_parameters}
\end{table}

\subsection{Baseline Models}
\phantomsection
\label{sec:baseline}
Our KA-GNN model has been evaluated alongside a selection of state-of-the-art GNN architectures, including both pre-trained and non-pre-trained models. These encompass MPNN frameworks like D-MPNN \cite{yang2019analyzing}, attention-driven models such as AttentiveFP \cite{xiong2019pushing}, and multi-scale approaches exemplified by Mol-GDL \cite{shen2023molecular}. Importantly, our comparisons also extend to geometry-focused graph models that leverage pre-training, including N-Gram \cite{liu2019n}, PretrainGNN \cite{hu2019strategies}, GROVER \cite{rong2020self}, GraphMVP \cite{liu2022pretraining}, MolCLR \cite{wang2022molecular}, GEM \cite{fang2022geometry}, Uni-mol \cite{zhou2023unimol}, SMPT \cite{LI2024108023}, and GNN-SKAN\cite{li2024}. Table \ref{tab:ka_gnn_gat_parameters} is the hyperparameters setting for various datasets, including model type, batch size, learning rate (LR), number of harmonics ($K$), number of message passing layers, number of attention heads (for GAT), and number of epochs.

\begin{table}[htbp]
\centering
\caption{Hyperparameter Sensitivity Analysis for KA-GNN.}
\begin{tabular}{lcccc}
\toprule
Batch size & 64 & 128 & 256 & 512  \\
\midrule
BACE & 0.887  & \textbf{0.890} & - & -  \\
BBBP & 0.696 & \textbf{0.787} & - & -  \\
ChinTox & \textbf{0.992}  &0.989& - & - \\
SIDER & 0.841 &\textbf{0.842} & - & -  \\
Tox21 & 0.772 & 0.774  & 0.769 &  \textbf{0.799} \\
HIV & 0.754  & 0.768  & 0.778  & \textbf{0.821}  \\
MUV &  0.686&0.696&0.725 & \textbf{0.834}   \\
\midrule
LR & 1e-3& 5e-4& 1e-4& 5e-5 \\
\midrule
BACE & 0.806  &0.818 & \textbf{0.890} &  0.858  \\
BBBP & 0.692 &0.717   & \textbf{0.787}& 0.736  \\
ChinTox & 0.984  &0.979   & \textbf{0.991}  & 0.989\\
SIDER &0.831   & 0.837 & \textbf{0.842} & 0.835   \\
Tox21 &0.725  & 0.763 & \textbf{0.799}& 0.764  \\
HIV &  0.754  & 0.779  & \textbf{0.821} & 0.819   \\
MUV &  0.708 & 0.723  & \textbf{0.834} &  0.801\\
\midrule
\#Layers & 0 & 1& 2& 3 \\
\midrule
BACE & 0.674 & 0.719 & 0.822 & \textbf{0.890}  \\
BBBP & 0.720 &  \textbf{0.787} & 0.734 & 0.672 \\
ChinTox & 0.973 &  0.976 & \textbf{0.989}&  0.984  \\
SIDER &  0.838& 0.830 & \textbf{0.842} &  0.829  \\
Tox21 & 0.702  &  0.721 & \textbf{0.799} & 0.760  \\
HIV & 0.718  & 0.763 & \textbf{0.821}&   0.754  \\
MUV &  0.700 &0.756   & \textbf{0.834} &    0.710\\
\bottomrule
\end{tabular}
\label{tab:parameters_sensitive}
\end{table}

\subsection{Hyperparameter Sensitivity Analysis}  
Table \ref{tab:parameters_sensitive} presents an analysis of the sensitivity of various hyperparameters, including batch size, learning rate (LR), and network depth, on the performance of molecular property prediction models. The results suggest that selecting an optimal batch size and a lower learning rate tends to enhance model performance. However, increasing the number of layers does not consistently yield improvements and may even result in overfitting or diminishing returns. These findings highlight the critical role of hyperparameter tuning in achieving optimal model performance.

\section{Conclusion} 
In this study, we propose the first non-trivial Kolmogorov-Arnold Network-based Graph Neural Networks (KA-GNNs). Our KA-GNNs utilize KAN’s unique power to optimize GNN architectures at three major levels, including node embedding, message passing, and readout. We develop Fourier series-based KAN model and provide a rigorous mathematical prove of the robust approximation capability of the Fourier KAN architecture. We compare the effects of various pre-activation function choices on the expressive power of the KAN model in terms of model performance and computational efficiency, identifying Fourier series as the optimal pre-activation function. Based on widely-used benchmark datasets for molecular property prediction, we extensively compare with existing state-of-the-art models, and find that our KA-GNNs can outperform traditional GNN models. This work not only highlights the great power of KA-GNNs in molecular property prediction but also provides a novel geometric deep learning framework for the general non-Euclidean data analysis.

\bibliographystyle{unsrt}  
\bibliography{references} 

\end{document}